\documentclass[conference]{IEEEtran}
\IEEEoverridecommandlockouts
\usepackage{cite}
\usepackage{amsmath,amssymb,amsfonts}
\usepackage{algorithmic}
\usepackage{graphicx}
\usepackage{textcomp}
\usepackage{xcolor}
\usepackage{url}
\def\BibTeX{{\rm B\kern-.05em{\sc i\kern-.025em b}\kern-.08em
    T\kern-.1667em\lower.7ex\hbox{E}\kern-.125emX}}
\begin{document}

\title{Mutual Information Maximization in Graph Neural Networks
}

\author{
\IEEEauthorblockN{Xinhan Di,
Pengqian Yu,
Rui Bu and
Mingchao Sun}
}

\maketitle

\begin{abstract}
A variety of graph neural networks (GNNs) frameworks for representation learning on graphs have been recently developed. These frameworks rely on aggregation and iteration scheme to learn the representation of nodes. However, information between nodes is inevitably lost in the scheme during learning. In order to reduce the loss, we extend the GNNs frameworks by exploring the aggregation and iteration scheme in the methodology of mutual information. We propose a new approach of enlarging the normal neighborhood in the aggregation of GNNs, which aims at maximizing mutual information. Based on a series of experiments conducted on several benchmark datasets, we show that the proposed approach improves the state-of-the-art performance for four types of graph tasks, including supervised and semi-supervised graph classification, graph link prediction and graph edge generation and classification. 
\end{abstract}

\begin{IEEEkeywords}
machine learning, neural networks, mutual information, graph theory, convolution
\end{IEEEkeywords}

\section{Introduction}
Learning with graph structure data requires effective representation of graph structure. Many approaches have been developed recently for this representation learning on graphs, such as graph convolutional networks (GCN) \cite{hamilton2017inductive,kipf2016semi,velivckovic2017graph,xu2018representation}, advanced pooling operation on graphs \cite{defferrard2016convolutional,simonovsky2017dynamic,ying2018hierarchical}, random walk-based methods \cite{grover2016node2vec,perozzi2014deepwalk,hamilton2017inductive}, mutual information neural estimation-based graph learning \cite{belghazi2018mine,velivckovic2018deep}, path-searching method \cite{shervashidze2009efficient}, tree-exploration strategy \cite{ramon2003expressivity}, and graph kernel framework \cite{rieck2019persistent}. Learned representation of the graph structure is then applied to graph tasks, including semi-supervised and supervised graph classification, graph link prediction, and graph embedding estimation.

Different neighborhood aggregation, graph-level pooling schemes, mutual information estimations, walk strategies and graph kernels are applied to GNNs \cite{hamilton2017representation} variants. These models have achieved state-of-the-art performance in a variety of tasks such as node classification, link prediction, and graph classification. However, the design of these new models has little analysis of the transformation of the adjacency matrix of the graph data. Besides, to the best of our knowledge, while complex approaches such as mutual information theory and expectation-maximization estimation yield good results, the current transformation of the adjacency matrix achieves little improvement on graph tasks. 

To further improve the performance of GNNs in the above tasks, we explore the scheme of aggregation and iteration of GNNs in the methodology of mutual information. In particular, we first introduce an equation of mutual information for the aggregation and iteration. We then update the equation of aggregation and iteration for the growth of mutual information. We further explore one of simple approaches aiming at the growth of mutual information. We implement this approach on several state-of-the-art graph models. According to the experimental results, performance improvements are achieved compared with a list of state-of-the-art graph models. Specifically, we obtain better performances for graph tasks such as supervised graph classification, semi-supervised graph classification, graph classification with missing edges and link prediction. Our implementation codes are available online at \url{https://github.com/CODE-SUBMIT/Graph_Neighborhood_1}.

\section{Related work}
There has been a rich line of research on graph learning models in recent years. Inspired by the first order graph Laplacian models \cite{kipf2016semi}, GCN is introduced to achieve promising performance on graph representation tasks, and its variants are proposed. For example, relational graph convolutional networks are developed for link prediction and entity classification \cite{scarselli2009graph}; linear mapping is applied to GCN in the concatenation module \cite{hamilton2017representation}; the sum module is developed in GCN for the aggregate representation of neighbors in the neighborhood \cite{xu2018powerful};  the capsule module is applied to the graph models \cite{verma2018graph} to solve especially graph classification problem; the knowledge network is used for the representation learning on graphs \cite{xu2018representation}, and hierarchical representation of graph is applied to different pooling strategies \cite{ying2018hierarchical}. However, these graph model variants rely on designs of complex convolution modules, and mutual information aspect of the aggregation and iteration scheme is not fully investigated. To fill in this research gap, we explore this aspect and propose an easily implementable approach for graph models in this paper.

Recent developed graph models have been combined with a variety of technologies in different fields. For instance, while the graph Markov neural network (GMNN) \cite{qu2019gmnn} is developed to combine Markov networks and graph convolutional networks (GMNN), the VAE module is parameterized with graph models based on an iterative graph refinement strategy (Graphite) \cite{grover2018graphite}. Inspired by the encoder-decoder architectures like U-Nets, both gPool and gunPool are developed to build the encoder-decoder model on graph (Graph U-Nets) \cite{gao2019graph}. There is also maximization of mutual information being applied to patch representation and corresponding high-level summaries of graphs \cite{velivckovic2018deep}, and a classical discrete probability distribution on the edges of the graph being learned for real problem while the given graph is incomplete or corrupted (LDS-GNN) \cite{franceschi2019learning}. However, it is unknown whether the information entropy plays an important role in the GNNs.

Many research efforts have been made on mutual information for graph neural networks \cite{xu2018powerful}. Information maximization is developed between edges states and transform parameters for the molecule property prediction tasks \cite{chen2019graph}. The mutual information is applied to maximize the mutual information between node representation and the pooled global graph representation \cite{velivckovic2017graph}. The student model is learned from the teacher model through the maximization of mutual information between intermediate representations learned by two models \cite{sun2019infograph}. However, to the best of our knowledge, the methodology of mutual information has not yet used for the analysis of the scheme of aggregation and iteration for GNNs. 

\section{Proposed Approach}

Graph neural networks generate node representations through aggregations over local node neighborhoods. GNN is a flexible class of embedding architectures and the representation of nodes are learned by aggregations of the neighborhood nodes. The READOUT function is used to summarize all the representations into a graph-level representation. The $k$-th layer of a GNN can be represented by
$
h_{v}^{k} = \text{COMBINE}^{k}(h_{v}^{k-1},\text{AGGREGATE}^{k}(\{(h_{v}^{k-1},h_{\mu}^{k-1},e_{\mu v}):\mu \in N(v)\}))
.$ Here node $v$ and nodes $\mu \in N(v)$, $h_{v}^{k}$ and $h_{\mu}^{k}$ are the corresponding feature vector at the $k$-th layer. $N(v)$ is the neighborhoods to node $v$.

\subsection{Improvement based on mutual Information}
During the learning of GNNs, we let the variable $V^{k}$, $M^{k}$ denote $h_{v}^{k}$, $M_{v}^{k}$ respectively. The above aggregation and iteration which depend on the variables such as $V^{k-1}$ and $M^{k-1}$ could be written as
\begin{equation}
I(V^{k}) \approx I(V^{k-1}) + \sum_{i=1}^{N}I(M_i^{k-1}), 
\end{equation}
where $M_{i}^{k-1}$ denotes the variable of $h_{\mu_{i}}^{k-1}$, $\forall, \mu_{i} \in N_(v), i\in\{1,...,N\}.$ The mutual information of $I(V^{k})$ depends on the sum of the above mutual information $I(V^{k-1})$ and $I(M_{i}^{k-1})$ where $i \in \{1,...,N\}$. 

The GNN is trained to learn the representation of each node (variable) by observing other nodes (variables). The mutual information is a measure of the amount of information about one variable through observing the other variables. In the framework of GNNs, the difficulty of learning the representation of nodes grows when observing more variables. In other words, the difficulty grows as the corresponding mutual information reduces.

Therefore, we update the above mutual information equation about the aggregation and iteration as follows
\begin{equation}
I(V^{k}) \approx I(V^{k-1}) + \sum_{i=1}^{N}I(M_i^{k-1}) + \sum_{j=1}^{N}I(N_{j}^{k-1}),
\end{equation}
where $N_{i}^{k-1}$ denotes the variable of $h_{\nu_{i}}^{k-1}$, $\forall, \nu_{j} \notin N_(v), j\in\{1,...,N\}, \nu \neq v.$

\subsection{Neighborhood enlargement for mutual information growth}
One simple way to implement the updated equation is to observe more nodes (variables). We propose an approach to enlarge the normal neighborhood in the aggregation of GNNs. The enlargement aims at a growth of mutual information for the scheme as analyzed above. The details are illustrated in the following.  

Let graph $G=(V,E)$ where each node $v \in V$ and each edge $e \in E$. Let $A$ denote the adjacency matrix of $G$, where a non-zero entry $A_{ij}$ indicates an edge between nodes $i$ and $j$. This adjacency matrix is a binary matrix ($A \in \{0,1\}^{n \times n}$). Let the length of all $e \in E$ be one. For $ v \in V$, the first neighborhood of $v$ consists of all vertices of distance one from $v$, denoted by $N_{1}(v)$. The complementary neighborhood of $v$ consists of all vertices of distance two from $v$, denoted by $N_{2}(v)$. Let $d(u,v)$ denote the distance between two vertices $u$ and $v$ where $u,v \in E$. The $d(u,v)$ is defined as the length of the path from the node $u$ to the node $v$. 

We let $A_{1}$ denote the adjacency matrix for the first neighborhood of the node $ v \in V$. That is, $A_{1_{ij}}=1,$ for  $v_{i}, v_{j} \in V, d(i,j)=1$ and $ A_{1_{ij}}=0$ if node $i$ and node $j$ are not adjacent. Let $A_{2}$ denote the adjacency matrix for the complementary neighborhood of the node  $v \in V$, i.e., $A_{2_{ij}}=1,$ for $ v_{i}, v_{j} \in V$ if  $d(i,j)=2$,  and $A_{1_{ij}}=0$ if $d(i,j) \neq 2$. Here $d(i,j)$ denotes the distance of the node $v_{i}$ and the node $v_{j}$. Note that $A_{1_{ij}}=A_{2_{ij}}=0$ for $i=j$. Besides, both $A_{1_{ij}}$ and $A_{2_{ij}}$ are the binary matrix.

In the literature, the adjacency matrix $A$ is equal to $A_{1}$. In contrast, we propose a transformation of the adjacency matrix as a combination of the above two neighborhoods $A=A_{1}+A_{2}$. This transformation is simple to implement and applicable to many state-of-art methods as we discuss below.

\subsection{Implementation of the neighborhood enlargement}

We first refer to the graph convolutional network (GCN) which is proposed by \cite{hamilton2017inductive}. This GCN graph convolutional layer is defined as:
\begin{equation}
    H^{i+1} = \sigma(A^{\wedge}H^{i}W^{i})
\end{equation}
where $H^{i} \in \mathcal{R}^{n \times s_{i}}$ and $H^{i+1} \in \mathcal{R}^{n \times s_{i+1}}$ are the input and output activations for layer $i$, $W_{i} \in \mathcal{R}^{s_{i} \times s_{i+1}}$ is a trainable weight matrix, $\sigma$ is an element-wise activation function, $A^{\wedge}$ is a symmetrically normalized adjacency matrix with self-connections, and $A^{\wedge}=D^{{-1}/{2}}(A+I_{n})D^{{-1}/{2}}$. Here $I_{n}$ is an $n \times n$ identity matrix. In this case, our proposed method has the following form
\begin{equation}
\begin{aligned}
    &H^{i+1} = \sigma(A_{s}^{\wedge}H^{i}W^{i}),\\
    &A_{s}^{\wedge} = D^{{-1}/{2}}(A_{1}+A_{2}+I_{n})D^{{-1}/{2}}.
    \end{aligned}
\end{equation}

Next, we refer to a discrete structure learning framework for graph neural network \cite{franceschi2019learning} where the graph structure and the parameters of graph convolutional networks are jointly learned. This jointly learning is based on a bilevel program given two objective functions $F$ and $L$: $F$ is the outer objectives for the learning of the outer function
\begin{equation}
    f_{w_{\theta}} = \textit{X}_{N} \times \textit{H}_{N} \rightarrow \textit{Y}^{N}
\end{equation}
and $L$ is the inner objectives for the learning of the inner function
\begin{equation}
    L(w_{\theta},A)=\sum_{v \in V_{train}}l(f_{w_{\theta}}(X,A)_{v},y_{v}),
\end{equation} 
where $f_{w}(X,A)_{v}$ is the output of $f_{w}$ for node $v$ and $l:\textit{Y} \times \textit{Y} \rightarrow \mathcal{R}^{+}$ is a point-wise loss function. Here $X \in \textit{X}_{N}$ is the feature matrix of the graph, $A \in \textit{H}_{N}$ is the adjacency matrix of G, $y \in \textit{Y}$ is the labels of the their true class,$ w \in \mathcal{R}^{d}$ and $\theta \in \mathcal{R}^{m}$ are the parameters of $f_{w_{\theta}}$. The bilevel program is then given by $\min \limits_{\theta,w_{\theta}}{F(w_{\theta},\theta)}$ such that $w_{\theta} \in \arg \min \limits_{w}{L(w,\theta)}$.

To apply our proposed transformation, we first replace $\textit{H}_{N}$ by $\textit{H}_{sN}$ where $H_{sN}=\{A_s| A_{s} = A+A_{2}, A\in H_N\}.$ For $ A_s \in \textit{H}_{sN}$, the outer and inner functions then become
\begin{equation}
\begin{aligned}
&f_{w_{\theta}} = \textit{X}_{N} \times \textit{H}_{sN} \rightarrow \textit{Y}^{N},\\
&L(w_{\theta},A_{s})=\sum_{v \in V_{train}}l(f_{w_{\theta}}(X,A_{s})_{v},y_{v}).
\end{aligned}
\end{equation}

Other state-of-the-art graph models such as GIN \cite{xu2018powerful}, GMNN \cite{qu2019gmnn}, Graph-Unets \cite{gao2019graph} and LDS-GNN \cite{franceschi2019learning} share a similar way of transformation.

\section{Experiments}

\begin{table*}[ht]
\caption{GCN and GIN supervised graph classification task on seven datasets.}\smallskip
\centering
\resizebox{1.85\columnwidth}{!}{
			\begin{tabular}{cccccccc}
			Graph models & MUTAG & 
				PROTEINS & PTC & NCL1 & IMDB-B & IMDB-M & COLLAB\\
			\hline
			GCN &$85.6\pm5.8$ &$76.0\pm3.2$ &$64.2\pm4.3$ &$80.2\pm2.0$ &$74.0\pm3.4$ &$51.9\pm3.8$ &$79.0\pm1.8$ \\
			(s)GCN &$86.23\pm6.71$ &$77.99\pm3.72$ &$69.23\pm5.59$ &$80.3\pm2.1$ &$76.1\pm2.86$ &$52.4\pm2.56$ & $80.89\pm2.3$\\ 
			GIN-$0$ &$89.4\pm5.6$ &$76.2\pm2.8$ &$64.6\pm7.0$ &$82.7\pm1.7$ &$75.1\pm5.1$ &$52.3\pm2.8$ &$80.2\pm1.9$ \\
			(s)GIN-$0$ &$94.14\pm2.74$ &$78.97\pm3.17$ &$73.56\pm4.27$ &$83.85\pm1.05$ &$77.94\pm4.31$ &$54.52\pm0.39$ & $80.71\pm1.48$\\
			GIN-$\epsilon$ &$89.0\pm6.0$ &$75.9\pm3.8$ &$63.7\pm8.2$ &$82.7\pm1.6$ &$74.3\pm5.1$ &$52.1\pm3.6$ &$80.1\pm1.9$ \\
			(s)GIN-$\epsilon$ &$93.47\pm1.64$ &$77.61\pm3.05$ &$72.16\pm2.17$ &$82.92\pm1.69$ &$75.19\pm5.1$ &$53.62\pm0.61$ & $80.51\pm1.62$\\
			\end{tabular}
			}
\label{table1}
\end{table*}

\begin{table*}[ht]
\caption{KNN-LDS supervised graph classification task on six datasets.}\smallskip
\centering
\resizebox{1.5\columnwidth}{!}{
			\begin{tabular}{ccccccc}
			Graph models & WINE & 
				CANCER & DIGITS & CITESEER & CORA & 20NEWS\\
			\hline
			KNN-LDS &$97.5\pm1.2$ &$94.9\pm0.5$ &$92.5\pm0.7$ &$71.5\pm1.1$ &$71.5\pm0.8$ &$46.4\pm1.6$\\
			(s)KNN-LDS &$98.0\pm1.1$ &$95.7\pm0.6$ &$92.5\pm0.6$ &$73.7\pm0.9$ &$72.3\pm0.6$ &$47.9\pm1.5$\\ 
			\end{tabular}
			}
\label{table2}
\end{table*}

\begin{table*}[ht]
\caption{GMNN semi-supervised graph classification task on three datasets.}\smallskip
\centering
    \resizebox{1.0\columnwidth}{!}{
			\begin{tabular}{cccc}
			Graph models & CORA & 
				CITESEER & PUBMED\\
			\hline
			GMNN &$83.4\pm0.8$ &$73.0 \pm 0.8$ &$81.3\pm0.5$\\
			(s)GMNN &$83.5\pm0.2$ &$73.4 \pm 0.1$ &$81.6\pm0.2$\\
			\end{tabular}
			}
\label{table4}
\end{table*}

\begin{table*}[ht]
\caption{Area under the ROC curve (AUC) for two baselines on link prediction task.}\smallskip
\centering
    \resizebox{1\columnwidth}{!}{
			\begin{tabular}{cccc}
			Graph models & CORA & 
				CITESEER & PUBMED\\
			\hline
			VGAE &$90.1\pm0.15$ &$92.0 \pm 0.17$ &$92.3\pm0.06$\\
			(s)VGAE &$93.4\pm0.12$ &$92.6 \pm 0.12$ &$92.7\pm0.05$\\
			Graphite-VAE &$91.5\pm0.15$ &$93.5.0 \pm 0.13$ &$94.6\pm0.04$\\
			(s)Graphite-VAE &$93.7\pm0.13$ &$94.1 \pm 0.10$ &$94.8\pm0.03$\\
			\end{tabular}
			}
\label{table5}
\end{table*}

\begin{table*}[!ht]
\caption{Average precision (AP) scores for two baselines on link prediction task.}\smallskip
\centering
    \resizebox{1\columnwidth}{!}{
			\begin{tabular}{cccc}
			Graph models & CORA & 
				CITESEER & PUBMED\\
			\hline
			VGAE &$92.3\pm0.12$ &$94.2 \pm 0.12$ &$94.2\pm0.04$\\
			(s)VGAE &$93.0\pm0.10$ &$94.3 \pm 0.08$ &$94.5\pm0.02$\\
			Graphite-VAE &$93.2\pm0.13$ &$95.0 \pm 0.10$ &$96.0\pm0.03$\\
			(s)Graphite-VAE &$93.5\pm0.11$ &$95.4 \pm 0.09$ &$96.3\pm0.02$\\
			\end{tabular}
			}
\label{table6}
\end{table*}

We conduct a bunch of experiments on various datasets and compare our proposed approach with the state-of-the-art graph models. In particular, our proposed method is applied to several graph learning tasks including supervised graph classification, semi-supervised graph classification, graph link prediction, and graph generation and classification with missing edges. The purpose is to show that our proposed approach outperforms the state-of-the-art methods for the listed datasets and is general to different GNN tasks. We also compare with other similar methods for adjacency matrix, and show that our proposed approach both works well with other methods and improves the performance of graph tasks. For a fair comparison, we keep all parameters the same as those in the baseline models GCN \cite{kipf2016semi}, GIN \cite{xu2018powerful},  KNN-LDS \cite{franceschi2019learning}, graph Markov neural network (GMNN) \cite{qu2019gmnn}, Graphite \cite{grover2018graphite}, Graph-Unet \cite{gao2019graph} and Mixhop \cite{abu2019mixhop}. In the following, (s)Model denotes our proposed model.

\begin{table*}[ht]
\caption{LDS on CITESEER and CORA with various percentage retained.}\smallskip
\centering
    \resizebox{1.4\columnwidth}{!}{
			\begin{tabular}{ccccc}
			Models (data) & 25\% & 50\% & 
				75\% & 100\% (full graph)\\
			\hline
			LDS (CITESEER) &$71.92\pm1.0$ &$73.26 \pm 0.6$ &$74.58\pm0.9$ &$75.54\pm0.4$\\
			(s)LDS (CITESEER) &$74.60\pm0.9$ &$74.90 \pm 0.7$ &$75.50\pm0.7$ &$76.11\pm0.6$\\
			LDS (CORA) &$74.18\pm1.0$ &$78.98 \pm 0.6$ &$81.54\pm0.9$&$84.08\pm0.4$\\
			(s)LDS (CORA) &$78.12\pm0.7$ &$80.41 \pm 0.7$ &$83.61\pm0.8$&$84.81\pm0.7$\\
			\end{tabular}
			}
\label{table7}
\end{table*}

\begin{table*}[ht]
\caption{GCN, MGCN, and MGCNK on ENZYMES with continuous node attribute and batch normalization.}\smallskip
\centering
    \resizebox{1.4\columnwidth}{!}{
			\begin{tabular}{cccc}
			\hline
				\multicolumn{4}{c}{ENZYMES dataset} \\
				\hline
			Models & - & Batch normalization & 
				Continuous node attribute \\
			\hline
			GCN &32.33$\pm5.07$ &$-$ &$51.17\pm 5.63$ \\
			(s)GCN &$40.11\pm3.27$ &$56.67\pm4.47$ &$75.33\pm3.93$ \\
			MGCN &$40.50\pm5.58$ &$-$ &$59.83\pm6.56$ \\
			(s)MGCN &$51.83\pm5.84$ &$59.23 \pm 4.12$ &$72.16\pm4.39$ \\
			MGCNK &$61.04\pm4.78$ &$-$ &$66.67\pm6.83$\\
			(s)MGCNK &$64.57\pm5.63$ &$65.12 \pm 4.28$ &$71.50\pm6.52$\\
			\end{tabular}
			}
\label{table8}
\end{table*}

\begin{table*}[ht]
\caption{GCN, MGCN, and MHCNK supervised graph classification task on two datasets without continuous node attribute and batch normalization.}\smallskip
\centering
   \resizebox{0.78\columnwidth}{!}{
			\begin{tabular}{ccc}
			Graph models & MUTAG & 
				PROTEINS\\
			\hline
			GCN &$76.5\pm1.4$ &$74.45 \pm 4.91$\\
			(s)GCN &$91.39\pm6.56$ &$78.49 \pm 3.19$\\
			MGCN &$84.4\pm1.6$ &$74.62 \pm 2.56$\\
			(s)MGCN &$87.25\pm4.16$& $78.23\pm3.23$\\
			MGCNK&$89.1\pm1.4$&$76.27\pm2.82$\\
			(s)MGCNK&$89.42\pm5.97$&$78.04\pm2.97$\\
			\end{tabular}
			}
\label{table9}
\end{table*}

\subsection{Supervised graph classification}
We first evaluate our approach on two state-of-the-art graph models GCN \cite{kipf2016semi} and GIN \cite{xu2018powerful}, which have achieved good results on both social and bi-logical graph datasets. Seven graph datasets are evaluated with our approach in comparison with these two baselines, including MUTAG (188 mutagenic aromatic and heteroaromatic nitro compounds with 7 discrete labels), PROTEINS (nodes for secondary structure elements and edge for neighbors in the amino-acid sequence or in 3D space), PTC (344 chemical compounds with 19 discrete labels), NCI1 (a dataset of chemical compounds with 37 discrete labels), IMDB-BINARY and IMDB-MULTI (movie collaboration datasets correspond to an ego-network for each actor/actress), and COLLAB (a scientific collaboration dataset derived from 3 public datasets). For a fair comparison, we keep all the parameters the same as the baseline models (GCN \cite{kipf2016semi} and GIN \cite{xu2018powerful}).

All experiments are conducted with 10-cross validation. Table \ref{table1} shows both average accuracy and standard deviation. Compared with GAN, our approach improves the average accuracy from $85.6\%$ to $86.23\%$ on MUTAG, from $76.0\%$ to $77.99\%$ on PROTEINS, from $64.2\%$ to $69.23\%$ on PTC, from $80.2\%$ to $80.3\%$ on NCI1, from $74.0\%$ to $76.1\%$ on IMDB-B, from $51.9\%$ to $52.4\%$ on IMDB-M and from $79.0\%$ to $80.89\%$ on COLLAB. Compared with GIN, our approach improves the average accuracy from $89.4\%$ to $94.14\%$ on MUTAG, from $76.2\%$ to $78.97\%$ on PROTEINS, from $64.6\%$ to $73.56\%$ on PTC, from $82.7\%$ to $83.85\%$ on NCI1, from $75.1\%$ to $77.94\%$ on IMDB-B, from $52.3\%$ to $54.52\%$ on IMDB-M and from $80.2\%$ to $80.71\%$ on COLLAB. It is clear that our proposed approach achieves better performance in comparison with the baselines. 

We then evaluate our approach on another state-of-the-art graph model KNN-LDS \cite{franceschi2019learning}. In this case, six graph datasets are evaluated with our approach and the baseline \cite{franceschi2019learning}. These datasets include WINE (a dataset of 178 samples with 3 discrete labels, each sample has 13 features), CANCER (a dataset of 569 samples with 2 discrete labels, each sample has 30 features), DIGITS (a dataset of 1797 samples with 10 discrete labels, each sample has 64 features), CITESEER (a data set of 3327 samples with 6 discrete labels, each sample has 3703 features), CORA (a dataset of 2708 samples with  7 discrete labels, each sample has 1433 features), 20NEWS (a dataset of 9607 samples with 10 discrete labels, each sample has 236 features). For a fair comparison, all parameters remain the same as the baseline \cite{franceschi2019learning}.

Similarly, all experiments are conducted with 10-cross validation. Both the average accuracy and standard deviation are presented in Table \ref{table2}. Our approach improves the average accuracy from $97.5\%$ to $98.0\%$ on WINE, from $94.9\%$ to $95.7\%$ on CANCER, from $71.5\%$ to $73.7\%$ on CITESEER, from $71.5\%$ to $72.3\%$ on CORA, and from $46.4\%$ to $47.9\%$ on 20NEWS. Our proposed approach achieves improvement in performance compared with the baseline.

\subsection{Semi-supervised graph classification}
In order to evaluate the effectiveness of our approach on semi-supervised graph classification, we apply the proposed transformation on the graph Markov neural network (GMNN) \cite{qu2019gmnn} which combines both statistical relational learning and graph neural networks. If the nodes have no labels, the neighbors of each node are predicted and the predicted neighbors are treated as pseudo labels. We apply our approach with neighborhood combination to evaluate the performance of the combined neighbors for the neighbor prediction and generation of pseudo labels. Three graph datasets are evaluated with our approach and GMNN including CORA, CITESEER, and PUBMED. In each object, we use the same data partition method as in the baseline \cite{qu2019gmnn}. Both our approach and GMNN are trained without the usage of object attributes since the difficulty of semi-classification drops with known object attributes.

All experiments are conducted with 10-cross validation, and both average accuracy and standard deviation are summarized in Table \ref{table4}. Our approach improves the average accuracy from $83.5\%$ to $83.4\%$ on CORA, from $73.0\%$ to $73.4\%$ on CITESEER, and from $81.3\%$ to $81.6\%$ on PUBMED. 

\subsection{Graph link prediction}
The task of link prediction is to predict whether an edge exists between a pair of nodes.  In order to evaluate how effective our proposed approach is for the link prediction, we apply the approach to the state-of-the-art graph models VGAE \cite{kipf2016variational} and Graphite \cite{grover2018graphite}. A balanced set of positive and negative (false) edges to the original graph are added to the original graph, with $5\%$ edges used for validation, and $10\%$ edges used for testing. Both the area under the ROC curve (AUC) and average precision (AP) metrics are assessed. Three datasets including CORA, CITESEER, and PUNMED are used for the evaluation.

All experiments are conducted with 10-cross validation. AUC is shown in Table \ref{table5} which shows the comparison between our approach and two baseline models: VGAE and Graphite-VAE. Compared with VGAE, our approach improves the AUC from $90.1\%$ to $93.4\%$ on CORA, from $92.0\%$ to $92.6\%$ on CITESEER, and from $92.3\%$ to $92.7\%$ on PUBMED. Our approach improves the AUC from $91.5\%$ to $93.7\%$ on CORA, from $93.5\%$ to $94.1\%$ on CITESEER, and from $94.6\%$ to $94.8\%$ on PUBMED when the base model is Graphite-VAE. The performance comparison can be found in Table \ref{table6}. Compared with VAE, our approach improves AP from $92.3\%$ to $93.0\%$ on CORA, from $94.2\%$ to $94.3\%$ on CITESEER, and from $94.2\%$ to $94.5\%$ on PUBMED. Our approach improves the AP from $93.2\%$ to $93.5\%$ on CORA, from $95.0\%$ to $95.4\%$ on CITESEER, and from $96.0\%$ to $96.3\%$ on PUBMED when the base model is Graphite.

\subsection{Graph classification with missing edges}
In practice, real-world graphs are often noisy and incomplete. As a consequence, the edges are usually missing in these graphs. In order to evaluate our approach for real-world graph tasks, we apply our approach to graph classification tasks with missing edges in the graphs. These graphs are obtained by randomly sampling $25\%$, $50\%$ and $75\%$ of the edges. Two datasets including CORA and CITESEER are evaluated. 

\begin{table*}[ht]
\caption{Supervised graph classification in comparison with two transformation forms and two baseline models.}\smallskip
\centering
   \resizebox{1.2\columnwidth}{!}{
			\begin{tabular}{cccc}
			GCN & GCN+$A^{2}$ & 
				GCN+$A^{2}$+$2I$ & (s)GCN\\
			\hline
			$74.37\pm0.31$ &$74.56\pm0.26$ &$74.23 \pm 0.37$&$78.49\pm 3.19$\\
            \hline
            Graph-Unet & Graph-Unet+$A^{2}$ & 
				Graph-Unet+$A^{2}$+$2I$ & (s)Graph-Unet\\
			$72.45\pm0.88$ &$72.87\pm0.52$ &$73.18 \pm 0.50$&$74.12\pm 4.17$\\
			\hline
			\end{tabular}
			}
\label{table10}
\end{table*}

\begin{table*}[ht]
\caption{Supervised graph classification for three datasets comparing Mixhop and (s)GMNN.}\smallskip
\centering
    \resizebox{1.0\columnwidth}{!}{
			\begin{tabular}{cccc}
			Graph models & CORA & 
				CITESEER & PUBMED\\
			\hline
			Mixhop &$71.4\pm0.81$ &$81.9 \pm 0.40$ &$80.8\pm0.58$\\
			(s)GMNN &$73.4\pm0.76$ &$83.5 \pm 0.57$ &$81.6\pm0.64$\\
			\end{tabular}
			}
\label{table11}
\end{table*}

All experiments are conducted with 10-cross validation, and both average accuracy and standard deviation are presented in Table \ref{table7}. Compared with LDS \cite{franceschi2019learning} for the dataset CITESEER, our approach improves the average accuracy from $71.92\%$ to $74.60\%$ when $75\%$ of the edges are missing, from $73.26\%$ to $74.90\%$ when $50\%$ of the edges are missing, from $74.58\%$ to $75.70\%$ when $25\%$ of the edges are missing, and from $75.54\%$ to $76.11\%$ when no edges are missing. Compared with LDS \cite{franceschi2019learning} for the dataset CORA, our approach improves the average accuracy from $74.18\%$ to $78.12\%$ when $75\%$ of the edges are missing, from $78.98\%$ to $80.41\%$ when $50\%$ of the edges are missing, from $81.54\%$ to $83.61\%$ when $25\%$ of the edges are missing, and from $84.08\%$ to $84.81\%$ when no edges are missing.

\subsection{Other transformations for adjacency matrix}
We also compare our approach with other state-of-the-art methods for the transformation of adjacency matrix. 
\subsubsection{Node attribute and batch normalization.}
Firstly, we compare our approach with three baseline graph models: GCN, MGCN, and MGCNK \cite{knyazev2018spectral} by applying continuous node attribute and batch-normalization. We also evaluate our approach by applying the spectral multigraph module. We compare our approach with the three graph models on dataset ENZYMES with continuous node attributes.

All experiments are conducted with 10-cross validation, and both average accuracy and standard deviation are shown in Table \ref{table8}. Applying our approach to GCN \cite{kipf2016semi} on ENZYMES leads to an increase in the average performance from $32.33\%$ to $40.11\%$, and from $51.17\%$ to $75.33\%$ with the usage of continuous node attribute and batch normalization. Compared with MGCN \cite{knyazev2018spectral} on ENZYMES, our approach improves the average performance from $40.50\%$ to $51.83\%$, from $59.83\%$ to $72.16\%$ with the usage of continuous node attribute and batch normalization. Compared with MGCNK \cite{knyazev2018spectral} on ENZYMES, our approach improves the average performance from $61.04\%$ to $64.57\%$, and from $66.67\%$ to $71.50\%$ with the usage of continuous node attribute and batch normalization.

\subsubsection{Multi-graph models}

Secondly, we evaluate our approach by applying two multi-graph modules and show its usefulness for multigraph models. Two datasets including MUTAG and PROTEINS are used. 

All experiments are conducted with 10-cross validation, and both average accuracy and standard deviation are reported in Table \ref{table9}. When the base model is GCN, our approach improves the average accuracy from $76.5\%$ to $91.39\%$ on MUTAG and from $74.45\%$ to $78.49\%$ on PROTEINS. Compared with MGCN \cite{knyazev2018spectral}, our approach improves the average accuracy from $84.4\%$ to $87.25\%$ on MUTAG and from $74.62\%$ to $78.23\%$ on PROTEINS. Compared with MGCNK \cite{knyazev2018spectral}, our approach improves the average accuracy from $89.1\%$ to $89.42\%$ on MUTAG and from $76.27\%$ to $78.04\%$ on PROTEINS.

\subsubsection{Powers of adjacency matrix and self-connectivity}

Thirdly, we evaluate our approach by applying the power of adjacency matrix or the addition of self-loops for adjacency matrix. Our approach is tested on the dataset PROTEINS. Its performance is compared against two baseline models with the application of the above transformation.

All experiments are conducted with 10-cross validation, and both the average accuracy and the standard deviation are shown in Table \ref{table10}. With the usage of $A^{2}$, the average accuracy increases from $74.37\%$ to $74.56\%$ on GCN \cite{hamilton2017inductive}, and from $72.45\%$ to $72.28\%$ on Graph-Unet \cite{gao2019graph}. With the usage of $A^{2}$+$2I$, the average accuracy increases to $74.23\%$ and $73.18\%$, respectively. Besides, our approach improves the accuracy from $74.37\%$ to $78.49\%$ on GCN and from $72.45\%$ to $74.12\%$ on Graph-Unet \cite{gao2019graph}.

\subsubsection{Convolutional transformation}

Finally, we test our approach with the transformation of adjacency matrix based on deep networks. The deep networks work on the calculation of the power of adjacency matrix. We compare our approach with the state-of-the-art Mixhop graph module \cite{abu2019mixhop}. Three datasets are used to evaluate the performances of different models.

All experiments are conducted with 10-cross validation, and both average accuracy and standard deviation are shown in Table \ref{table11}. Our approach, which applies to GMNN \cite{qu2019gmnn}, achieves better performance on all three datasets: it achieves $73.4\%$ on CORA while Mixhop obtains $71.4\%$; it achieves $83.5\%$ on CITESEER while Mixhop achieves $81.9\%$, and it achieves $81.6\%$ on PUBMED while Mixhop \cite{abu2019mixhop} only reaches $80.8\%$.

\section{Discussion}
In this paper, we analyze the scheme of the aggregation and iteration in the graph neural networks in the methodology of mutual information. We propose an approach to enlarge the normal neighborhood in the aggregation of GNNs. We apply the proposed method on several of the state-of-the-art graph neural network models and conduct a series of experiments on the graph representation tasks including supervised graph classification, semi-supervised graph classification, and graph edge generation and classification. The numerical results on various datasets show that our approach improves the performance of the listed state-of-the-art models. For the future work, it is worthwhile to study new network architecture, message transmission and graph kernels in the proposed approach, and test on large-scale graph tasks.

{\bibliographystyle{ieeetr}\bibliography{bib1}}

\end{document}